\documentclass[journal]{IEEEtran}
\usepackage[utf8]{inputenc}
\usepackage[T1]{fontenc}
\usepackage{graphicx}
\usepackage{wrapfig}
\usepackage{float}
\graphicspath{ {./images/} }
\title{\LARGE \bf
Global Sentiment Analysis Of COVID-19 Tweets Over Time*
}

\author{Muvazima Mansoor$^{1}$, Kirthika Gurumurthy$^{2}$, Anantharam R U$^{3}$, and V R Badri Prasad$^{4}$
\thanks{*This work was not supported by any organization}
\thanks{$^{1}$Muvazima Mansoor is with the Department of Electronics and Communication,
        PES University, Bengaluru, Karnataka, India
        {\tt\small muvazima99@gmail.com}}%
\thanks{$^{2}$Kirthika Gurumurthy is with the Department of Computer Science,
        PES University, Bengaluru, Karnataka, India
        {\tt\small kirthikagurumurthy@gmail.com}}%
\thanks{$^{3}$Anantharam R U is with the Department of Computer Science,
        PES University, Bengaluru, Karnataka, India
        {\tt\small anantharamru99@gmail.com}}%
\thanks{$^{4}$Badri Prasad is with the Faculty of Department of Computer Science,
        PES University, Bengaluru, Karnataka, India
        {\tt\small badriprasad@pes.edu}}%

}

\begin{document}

\maketitle
\thispagestyle{empty}
\pagestyle{empty}

\begin{abstract}

The Coronavirus pandemic has affected the normal course of life. People around the world have taken to social media to express their opinions and general emotions regarding this phenomenon that has taken over the world by storm.
The social networking site, Twitter showed an unprecedented increase in tweets related to the novel Coronavirus in a very short span of time. This paper presents the global sentiment analysis of tweets related to Coronavirus and how the sentiment of people in different countries has changed over time. Furthermore, to determine the impact of Coronavirus on daily aspects of life, tweets related to Work From Home (WFH) and Online Learning were scraped and the change in sentiment over time was observed. In addition, various Machine Learning models such as Long Short Term Memory (LSTM) and Artificial Neural Networks (ANN) were implemented for sentiment classification and their accuracies were determined. Exploratory data analysis was also performed for a dataset providing information about the number of confirmed cases on a per-day basis in a few of the worst-hit countries to provide a comparison between the change in sentiment with the change in cases since the start of this pandemic till June 2020.
\end{abstract}

\section{INTRODUCTION}
The pandemic has had a pronounced effect on the lives of human beings across the world. The outbreak has not only created havoc in physical health,
economic conditions, working conditions, and the manufacturing sector to name a few but has also created a niche in the minds of the people around the world. It has had serious repercussions on the psychological state of the humans that is most evident now [1]. Social Media platforms like Twitter are a great resource to capture human emotions and thoughts. During these trying times, people have taken to social media to discuss their fears, opinions, and insights on the global pandemic. This paper is an effort towards the analysis of tweets and the change in sentiments of the people globally from January 2020 to June 2020. Due to government-imposed lock-downs to contain the spread of the virus, people were forced to adapt to Work From Home scenarios, and educational institutions were asked to implement online learning. Some people might regard WFH as a necessary step, while others regard it as an inconvenience due to the lack of high-speed internet or smart devices. This paper presents how some daily aspects of life have been affected by analyzing people's sentiment towards online learning and WFH scenarios. 
To gauge the sentimental content of the tweets, a technique called sentiment analysis was used to label the tweets as Positive, Negative, or Neutral. A  dataset created from scraping tweets and labeling the sentiment using the sentiment lexicon VADER was then used to train classification models such as LSTM and ANN. While Sentiment Analysis of the tweets can reflect the general opinion of people, it does not convey the actual impact of the virus that is seen in the number of cases and deaths that are exponentially increasing every day. Therefore, exploratory data analysis of a dataset containing the daily number of cases and deaths in a few of the worst-hit countries was performed as well to provide a comparison between the public sentiment and the rise or fall in cases globally.

\section{LITERATURE REVIEW}
This study was informed by multiple research articles from various disciplines and therefore, in this section literature review on sentiment analysis, Vader, LSTM, and ANN are covered. 
\subsection{Sentiment Analysis}
Munikar et al [2] describe Sentiment analysis as a supervised machine learning problem. There are different types of sentiment analysis including fine-grained sentiment analysis, aspect-based sentiment analysis, and emotion detection. In binary sentiment classification, the possible classifications are positive and negative. In fine-grained sentiment classification, there are five groups- positive, very positive, neutral, negative, and very negative. In this paper, we have performed binary sentiment classification and emotion detection, primarily fear and trust.

\subsection{Vader}
Gilbert [3] developed VADER, which is a rule-based model for general sentiment analysis, and compared its effectiveness to 11 typical benchmarks, including Word Count (LIWC), Affective Norms for English Words(ANEW), the General Inquirer, Linguistic Inquiry, Senti WordNet, and machine learning techniques that rely on Support Vector Machine (SVM) algorithms, Naive Bayes and, Maximum Entropy. The study described the validation, evaluation, and development of VADER. The author used a combination of qualitative and quantitative methods to produce a sentiment lexicon that is used in the social media domain. VADER uses a parsimonious rule-based model to gauge the sentiment of tweets. 

\subsection{LSTM Neural Networks}
Hochreiter et al.[4] analyze the vanishing gradient problem in detail.
When the gradient of the error function of the neural network is propagated back through a unit of a neural network, it
gets scaled by a factor smaller than one or greater than one.
Due to this, the gradient blows up or decays exponentially over time in a recurrent neural network. Therefore, the gradient either gets lost or dominates the next weight adaptation step. To overcome this shortcoming,
the authors re-designed the unit
of a neural network in such a way that the scaling factor is always one. To overcome the limitation in the learning capabilities due to the new unit, it was enriched by gating units. The gating units add the activations of the current layer and the activations of the previously hidden layer from the previous time step and the inner activation of the LSTM unit. The LSTM unit along with the gating units can be interpreted as a differentiable version of computer memory. Thus, LSTM units are also known as
LSTM memory cells. Sundermeyer et al.[5] mention that the LSTM architecture solves the vanishing gradient problem at small computational extra-costs. It also has the desirable property of including standard recurrent neural network units as a special case.

\subsection{Artificial Neural Networks}
James [6] describes the Artificial Neural Network as a collection of processing elements that are interconnected and transform a set of inputs into a set of desired outputs. The result of the transformation is determined by the characteristics of the elements and the weights associated with the interconnections among them. By modifying the connections between the nodes the network is able to adapt to the desired outputs. A neural network analyzes the information and provides a probability estimate that the data matches the characteristics which it has been trained to recognize. The accuracy decisions by the neural network rely totally on the experience the system gains in analyzing instances of the stated problem. The neural network gains experience by training the system to correctly identify the preselected instances of the problem. The response of the neural network is analyzed and the configuration of the system is corrected until the neural network’s analysis of the training data reaches a satisfactory level. The neural network gains experience over time as it performs analyses on data related to the problem.

\subsection{Related Works}

Rajput et al.[1] present a statistical analysis of the twitter messages related to Coronavirus posted since January 2020. They performed two types of empirical studies. The first on word frequency and the second on sentiments of the individual tweet messages. Unigram, bigram, and trigram frequencies were modeled by a power-law distribution. The results were validated by the Sum of Square Error (SSE), R2, and Root Mean Square Error (RMSE). High values of R2 and low values of SSE and RMSE lay the grounds
for the goodness of fit of the model. The results showed that the majority of the tweets had a positive polarity and only about 15\% were negative.
\newline
Samuel, Jim, et al.[7] identify the general public sentiment related to the pandemic using Coronavirus specific Tweets using R statistical software and its sentiment analysis packages. The authors demonstrate an insight into the progress of fear-sentiment over time as COVID-19 approached peak levels in the US, using descriptive textual analytics. In addition, they provide an overview of two essential machine learning (ML) classification methods-Naive Bayes and Logistic Regression, and compare their effectiveness in classifying tweets of varying lengths. Naive Bayes gave an accuracy of 0.9143 for shorter tweets and an accuracy of 0.5714 for longer tweets. Whereas, Logistic Regression gave an accuracy of 0.7429 for shorter tweets and an accuracy of 0.52 for longer tweets.
\newline
Barkur et al. [8], analyze sentiments of Indians regarding lockdown announcements. The authors extracted Tweets using two prominent hashtags namely: \#IndiaLockdown and \#IndiafightsCorona from March 25th to March 28th, 2020. A total of 24,000 tweets were considered for the analysis done using R statistical software. The prominent sentiment was positive, followed by the trust. Overall, the results show that Indians have taken the fight against Covid19 positively and the majority are in agreement with the government for announcing the lockdown to flatten the curve.
\newline
Li, Sijia, et al[9] explore the impacts of COVID-19 on people’s mental health, assist policymakers to develop actionable policies, and help clinical practitioners provide timely services to affected populations. The authors analyze and sample Weibo posts from 17,865 active users using the approach of Online Ecological Recognition based on several machine-learning predictive models. They calculated word frequency, scores of emotional indicators, and cognitive indicators from the collected data. The paired sample t-test and sentiment analysis were performed to examine the differences in the same group before and after the declaration of COVID-19 on 20 January 2020. The results showed that sensitivity and negative emotions towards social risks increased, while the scores of life satisfaction and positive emotions decreased.
\newline
Jang, Hyeju, et al.[10] investigate people's reactions and concerns about COVID-19 in North America, especially focusing on Canada. They analyze COVID-19 related tweets using topic modeling and aspect-based sentiment analysis and interpret the results with public health experts. They compare the timeline of topics discussed with the timing of implementation of public health interventions for COVID-19. They also examine people's sentiment about COVID-19 related issues. Finally, they discuss how the results can be helpful for public health agencies when designing a policy for new interventions.
\newline
Zhou, Jianlong, et al.[11] used tweets to analyze the sentiment dynamics of people living in the state of New South Wales (NSW) in Australia during the pandemic period. In addition, this study analyzed the sentimental dynamics delivered by the trending topics on Twitter such as government policies (e.g. lock-down, social-distancing, Australia's JobKeeper program) as well as the focused social events like the Ruby Princess Cruise.

\section{Datasets}
\subsection{Coronavirus Tweets Dataset}
165116 tweets were scraped based on the keywords “coronavirus”  using  TwitterScraper[12] which uses the python package requests to retrieve the content and Beautifulsoup4 to parse the retrieved content. Unlike the traditionally used twitter API,  it is not limited to scraping tweets for only a duration of 7 days. These tweets have been fetched from all around the world from Jan 1st, 2020 to Jun 29th, 2020. To extract the countries from which the tweets originated, a geolocation filter was specified. However, since most Twitter users do not have geotagging enabled for tweets, the location was extracted from the profile of the username of the tweet using the Tweepy Python library. To extract the country from the location, a world city dataset [13] from datahub.io was used which consisted of the mapping of cities all around the world to their corresponding country, sub country, geoname ID, and abbreviation. If the location extracted from the user’s profile matched any of the country, sub country, geoname ID, city, or abbreviation columns, the tweet was assigned to the corresponding country. Next, to obtain the sentiment of these tweets the rule-based sentiment analysis tool VADER [14] was used, which is specifically attuned to sentiments expressed in social media. The tweets were assigned positive, neutral, and negative sentiments based on their VADER compound score. The columns present in this created dataset include the tweet text, date, country, and sentiment.

\subsection{ Online Learning Dataset (Tweets)}
40756 tweets were scraped using the keywords “online learning“ and “online classes” with the help of Twitterscraper[12] from 19th Feb 2020 to 26th Jun 2020.The method employed to extract the countries and the sentiments of tweets were the same as that of the coronavirus tweets dataset. The columns present in this dataset include the tweet text, date, country, and sentiment.
\subsection{Work from Home Dataset (Tweets)}
41349 tweets were scraped using the queries “work from home” and "WFH” with the aid of TwitterScraper[12] for the duration 22nd Feb 2020 to 3rd May. The countries were extracted using the world cities dataset as described previously and the tweets were assigned sentiments with the help of the VADER sentiment lexicon. The columns present in this dataset include the tweet text, date, country, and sentiment.

\subsection{ Existing dataset usage : Covid19 Dataset}
A Covid-19 dataset [15] obtained from Kaggle was used which consists of information pertaining to the number of confirmed, death, and recovered cases every day across the globe. It contains data from 22-01-2020 to 22-06-2020.

\section{Methodology}
\subsection{Exploratory Data Analysis}
\subsubsection{Binary Sentiment}
The process of exploratory data analysis started with the visualization of the sentiments of the people around the world towards coronavirus, work from home and online learning. To plot the world map, a new dataset containing the count of positive and negative tweets for each country was created. The count of positive and negative tweets were normalized and plotly was used to map the country and the sentiment count on the graphs shown below. (fig.1-fig.6).
\begin{figure}[H]
      \centering
      \includegraphics[scale=0.25]{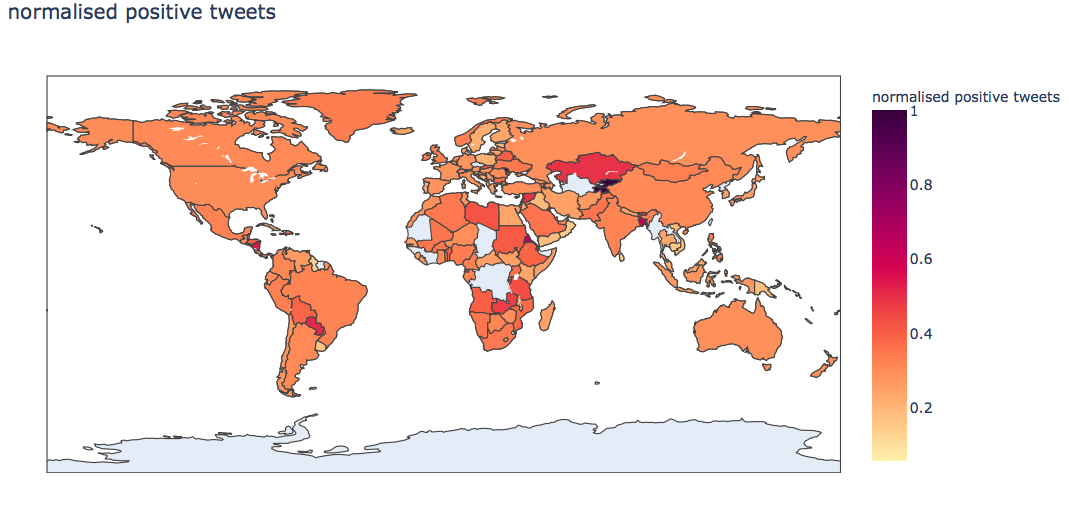}
      \caption{Positive sentiment towards coronavirus}
      \label{figurelabel1}
   \end{figure}
\begin{figure}[H]
      \centering
      \includegraphics[scale=0.25]{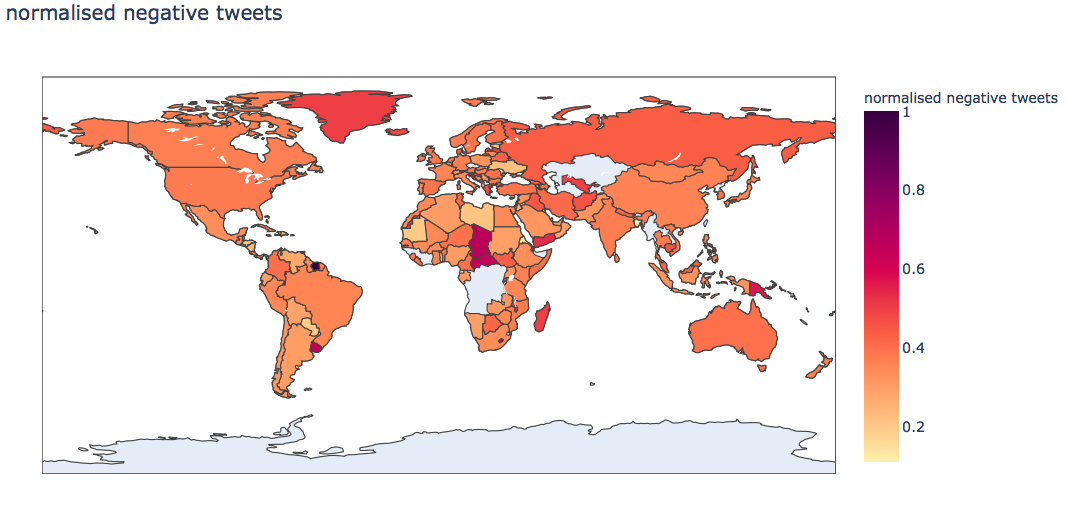}
      \caption{Negative sentiment towards coronavirus}
      \label{figurelabel2}
   \end{figure}
\begin{figure}[H]
      \centering
      \includegraphics[scale=0.25]{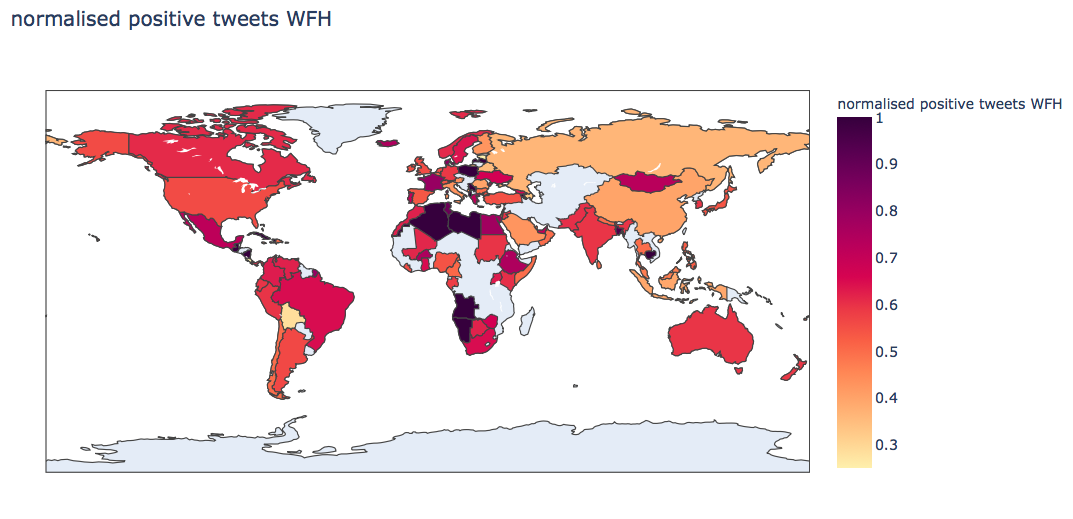}
      \caption{Positive sentiment towards Work From Home}
      \label{figurelabel3}
   \end{figure}   
\begin{figure}[H]
      \centering
      \includegraphics[scale=0.25]{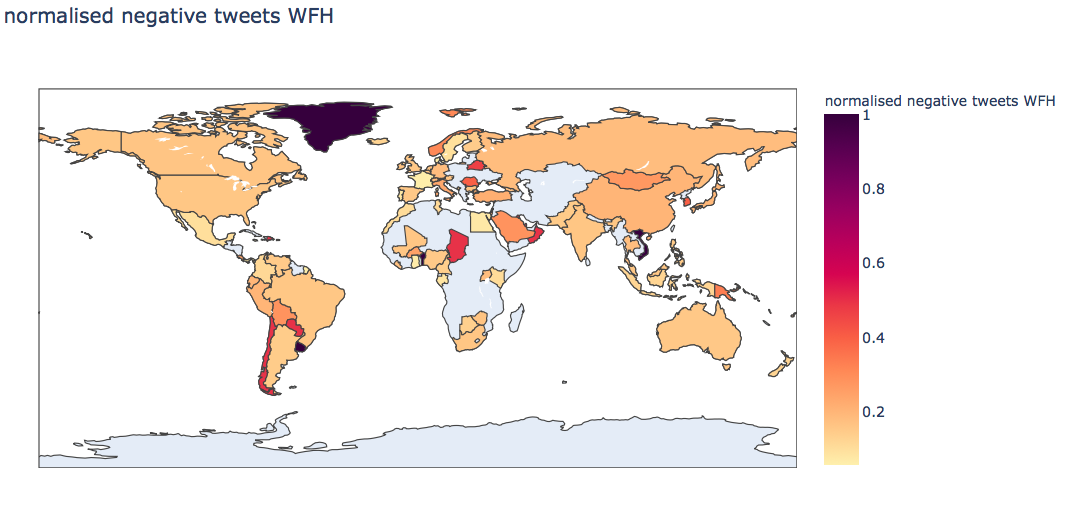}
      \caption{Negative sentiment towards Work From Home}
      \label{figurelabel4}
   \end{figure}
\begin{figure}[H]
      \centering
      \includegraphics[scale=0.25]{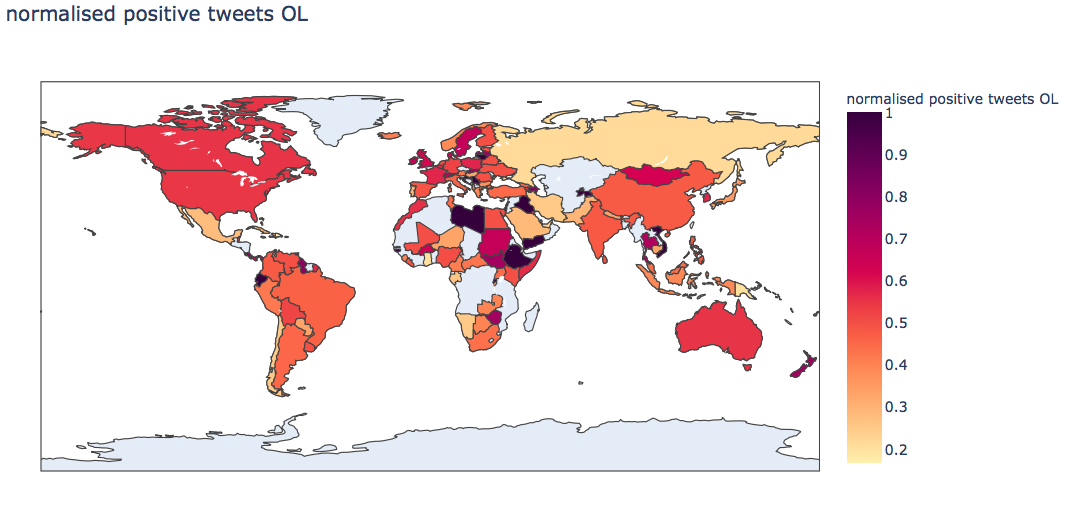}
      \caption{Positive sentiment towards Online Learning}
      \label{figurelabel5}
   \end{figure}   
\begin{figure}[H]
      \centering
      \includegraphics[scale=0.25]{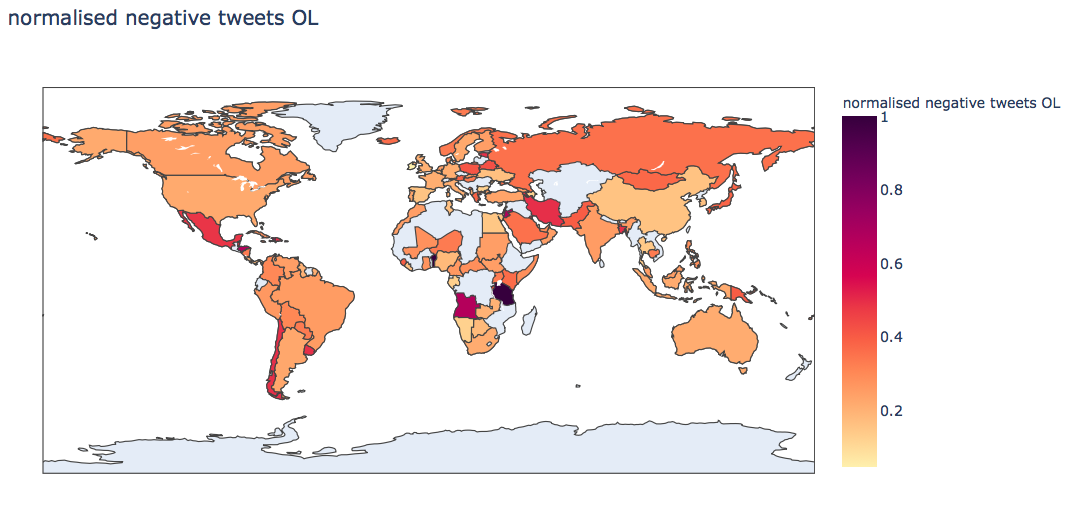}
      \caption{Negative sentiment towards Online Learning}
      \label{figurelabel6}
   \end{figure}
For a better understanding of the change in sentiment over time, the tweets were grouped by month and the percentage of positive and negative tweets in these months were plotted. In addition to plotting the general sentiment around the world, the analysis was also done on the worst hit countries like- United States of America, Brazil and India.
To obtain a more fine grained analysis, a time series graph of the sentiment vs time was also plotted for the entire world and the countries mentioned above. 
Since the sentiment of the people is not necessarily related to the number of cases, the countries with the most positive and negative sentiments towards coronavirus, work from home and online learning were found.

\subsubsection{Emotion Analysis}
 The analysis was done for the  Coronavirus tweets dataset with respect to the fear and trust emotions exhibited by the tweets. For this, the NRC emotion lexicon [16] aka EmoLex was used, which is a list of words and their associations with eight basic emotions (anger, fear, anticipation, trust and surprise, sadness, joy, and disgust). Using the python NRCLex library raw emotion scores were found, which provide the count of the number of words in the text associated with each particular emotion. The final emotion score was then calculated by dividing the count of each emotion by the total count of all the emotions. The predominant emotion of the tweet was then retrieved by finding the emotion having the maximum score. For this study, we take into consideration only the emotions of fear and trust exhibited by the tweets.
\newline
A bar graph was plotted to visualize the percentage of tweets denoting fear and trust as their primary emotion for each month. A time series was also plotted on a day-wise basis to see the trend in the change in these emotions. This was also done separately for the countries that have been worst hit by the pandemic such as the US, India, and Brazil. Finally, the countries having the highest fear and trust scores were obtained for a comparison between the emotion and the number of cases in those countries (the analysis of which is described in the following section).

\subsubsection{Visualisation of the number of confirmed cases }
An exploratory data analysis on the number of confirmed cases was carried out as per the data provided in the Kaggle covid19 cases dataset to provide a comparison of the same with the public sentiment and emotions. A time series was plotted to visualize the change in the number of cases from January 22nd, 2020 to June 22nd, 2020. The number of cases per month was plotted as a bar graph. The same visualizations were obtained for the countries - US, India, and Brazil.

\subsection{Machine Learning with Classification Methods}
Recently, deep learning-based methods for sentiment classification (such as LSTM, ANN, etc) have been gaining popularity due to impressive results depicted by the same. The polarity of the opinions expressed by the tweets has been determined using two deep learning models - Artificial Neural Network (ANN) and Long Short Term Memory (LSTM). 

\subsubsection{ANN}
A multi-layer perceptron (MLP) classifier was trained on the dataset. MLPClassifier implements an MLP algorithm that trains using Backpropagation. It is part of the sklearn estimator class in the scikit-learn module of Python. The input to the classifier was word vectors of each review which was created using the bag of words model. The activation function used was Relu and the Weight optimization was done using a stochastic gradient-based optimizer called 'adam' proposed by Kingma, Diederik, and Jimmy Ba[17]. This solver is more suitable for large datasets. A total of 2000 neurons were given in the hidden layer and a standard learning rate of 0.001 was used. 

\subsubsection{LSTM}
After performing the basic preprocessing of the tweets (such as removing punctuations, converting to lowercase), a vocabulary size was defined to convert the words into vectors. The labels were encoded as well and padding/truncating was done to bring all the vectors to a specific length. The LSTM network architecture included the following layers - the Embedding layer (used to create word vectors for the incoming words), the LSTM layer, the Dropout layers for avoiding overfitting, the Conv1D layer to summarize the vector along one dimension, the MaxPooling1D layer that calculates the maximum, or largest, the value in each patch of each feature map and finally the last layer as a fully connected layer with a ‘softmax’ activation function and neurons equal to the number of unique categories (in this case : 3 ) to obtain a one-hot encoded result. 

\section{Results}
Time-series graphs of the number of cases, binary sentiment (positive and negative), fear and trust emotions with respect to coronavirus, work from home, and online learning were plotted and analyzed for the entire world as well as the worst-hit countries like the USA, India, and Brazil. The accuracies of the two deep learning models employed for sentiment classification were obtained. 

\subsection{Accuracy of Machine Learning models}
On the coronavirus tweets dataset labeled using VADER, an accuracy of 84.5\% was obtained using LSTM and an accuracy of 76\% was obtained using ANN.

\subsection{Worldwide}
\begin{figure}[H]

      \centering
      \includegraphics[scale=0.2]{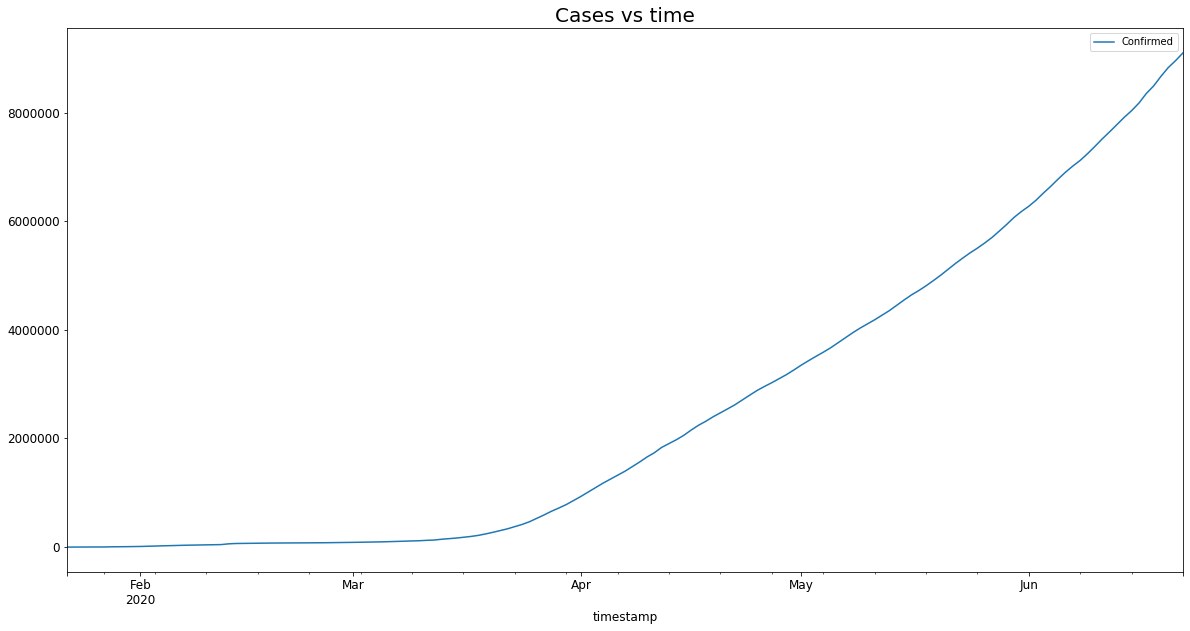}
      \caption{Number of cases vs time}
      \label{figurelabel7}
   \end{figure}   
The number of cases saw an exponential increase in the months after March as seen in Fig 7.
\begin{figure}[H]
      \centering
      \includegraphics[scale=0.2]{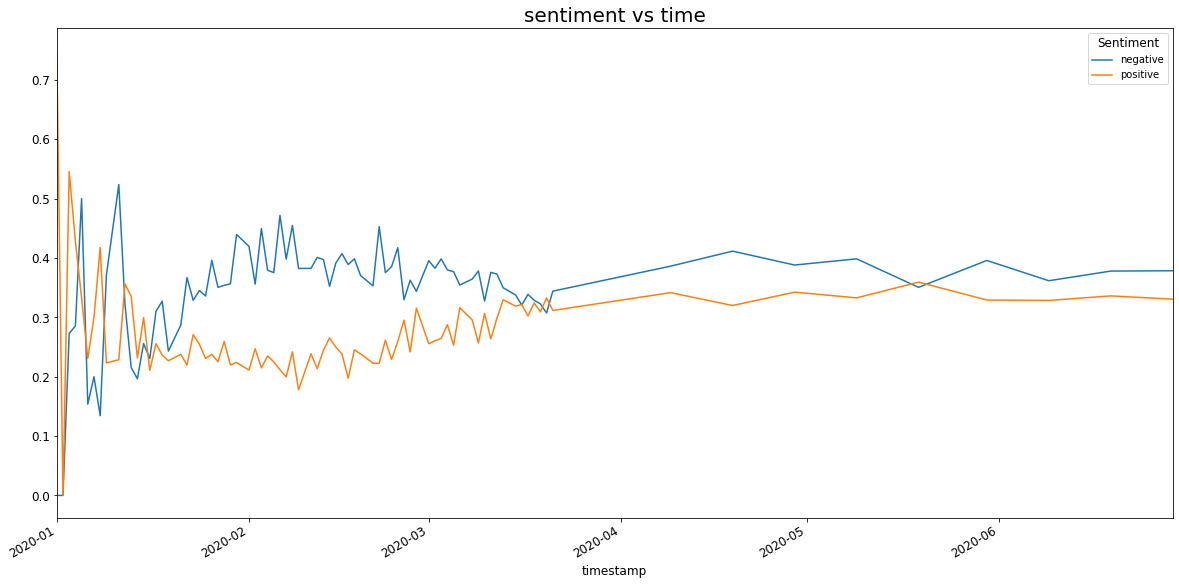}
      \caption{Binary sentiment vs time}
      \label{figurelabel8}
   \end{figure}  
In Fig 8., it is observed that the gap between the percentage of negative and positive sentiments was much greater in February and March as compared to the subsequent months.
\begin{figure}[H]
      \centering
      \includegraphics[scale=0.2]{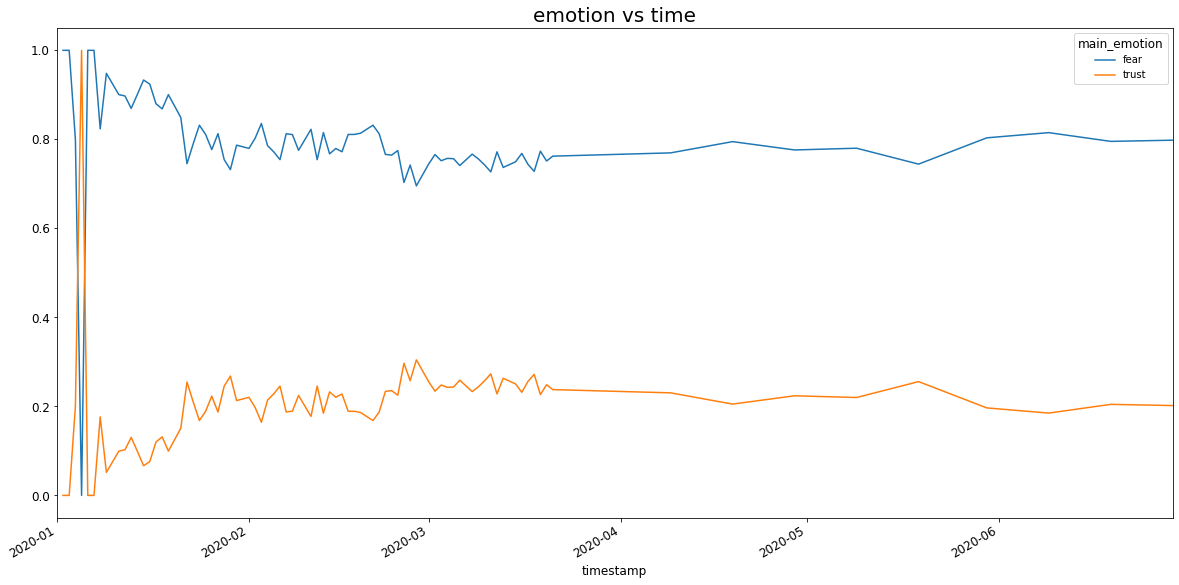}
      \caption{Fear and Trust sentiment vs time}
      \label{figurelabel9}
   \end{figure}
Fig. 9 illustrates that the Fear emotion has been significantly greater than the trust emotion throughout the pandemic.
Some of the countries with a greater proportion of positive sentiment are Bangladesh, Pakistan, Mali, and South Africa. Whereas, Australia, India, Canada, the USA, Turkey, UK, and Brazil are some of the countries where the proportion of negative sentiment is higher.
The emotions score calculated using the NRC lexicon (as described in the  Methodology section) showed the following results - Thailand, Vietnam, and Poland were found to have the highest fear score. Meanwhile, the highest trust scores were exhibited by Oman, Syria, and Kazakhstan.
  
\subsection{Worst Hit Countries}
\subsubsection{USA}
The number of cases in US saw an exponential increase in the months after March as seen in Fig 10.   
\begin{figure}[H]
      \centering
      \includegraphics[scale=0.2]{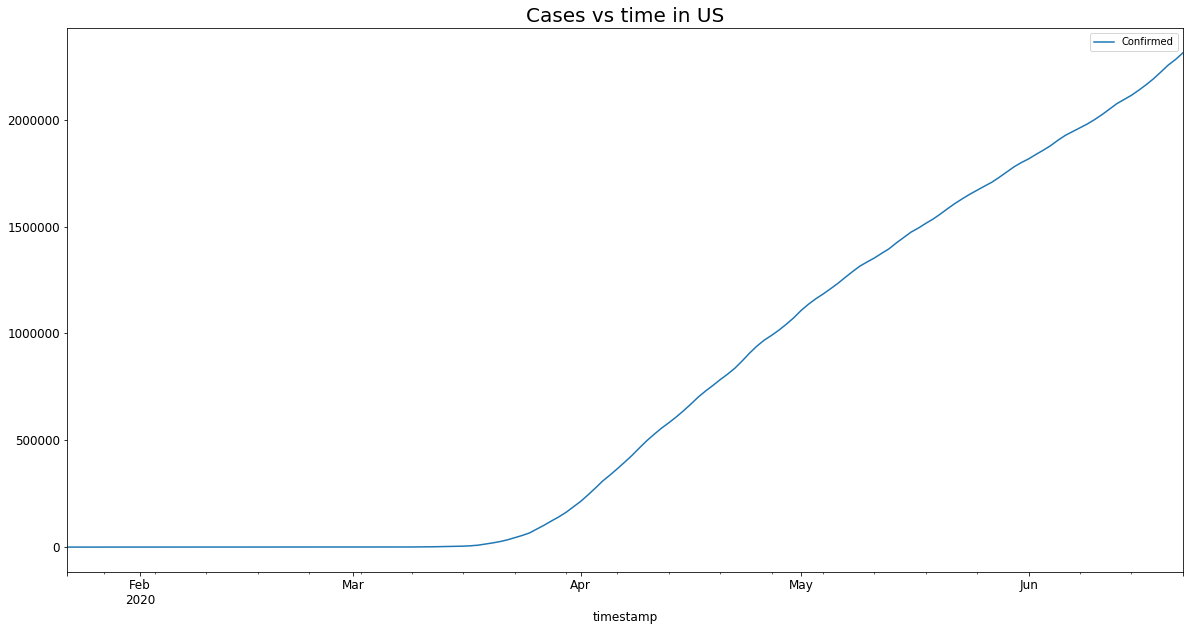}
      \caption{Number of cases in US vs time}
      \label{figurelabel10}
   \end{figure}

\begin{figure}[H]
      \centering
      \includegraphics[scale=0.25]{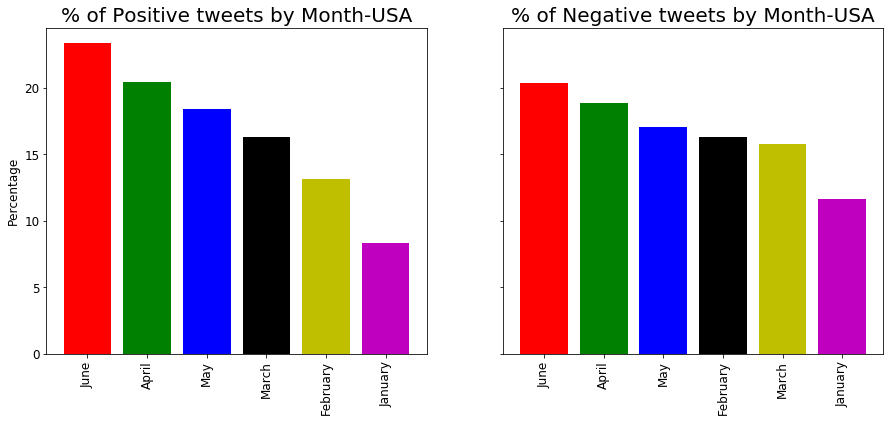}
      \caption{Percentage of positive and negative sentiment in US per month}
      \label{figurelabel11}
   \end{figure}

\begin{figure}[H]
      \centering
      \includegraphics[scale=0.25]{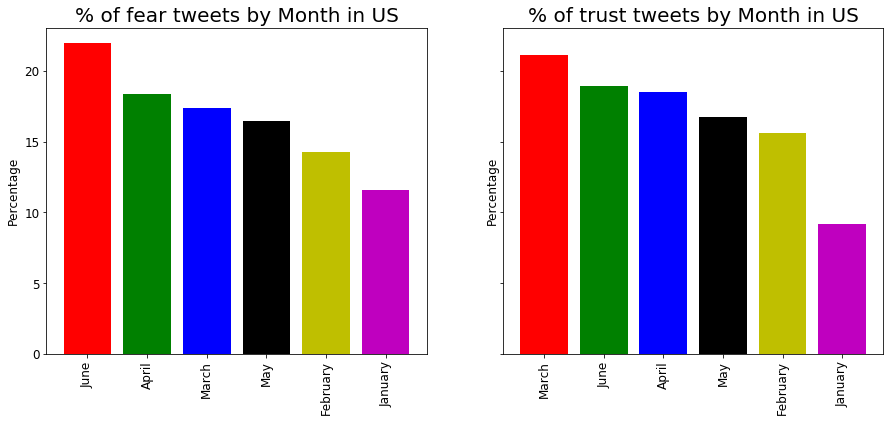}
      \caption{Percentage of Fear and Trust emotion in US per month}
      \label{figurelabel12}
   \end{figure}   
The greater percentage of trust emotion in the month of March as seen in Fig 12. could be attributed to the introduction of lockdowns imposed by the government to contain the virus. The greater percentage of fear emotion in the month of April could be a result of the exponential growth in the number of cases in April.
\subsubsection{India}
The number of cases in India saw an exponential increase in the months after April as seen in Fig 13.
\begin{figure}[H]
      \centering
      \includegraphics[scale=0.2]{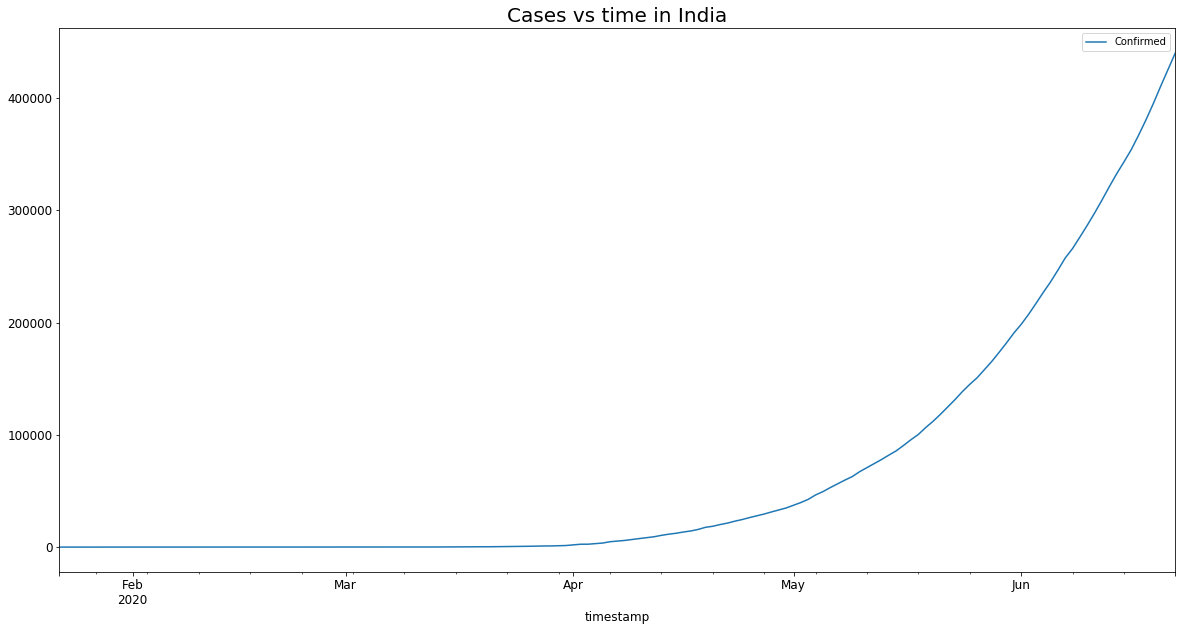}
      \caption{Number of cases in India vs time}
      \label{figurelabel13}
   \end{figure}  

\begin{figure}[H]
      \centering
      \includegraphics[scale=0.25]{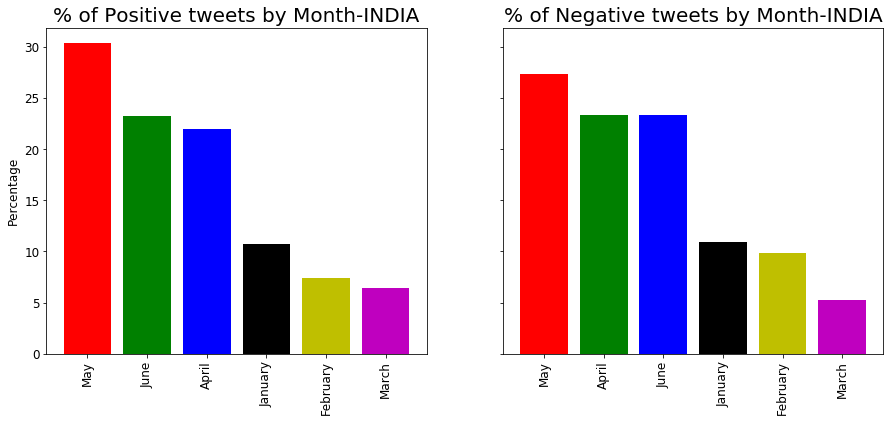}
      \caption{Percentage of positive and negative sentiment in India per month}
      \label{figurelabel14}
   \end{figure}
An increase in positive and negative tweets in May as shown in Fig. 14 can be attributed to the exponential increase in the number of cases in May in India. 
 
\begin{figure}[H]
      \centering
      \includegraphics[scale=0.25]{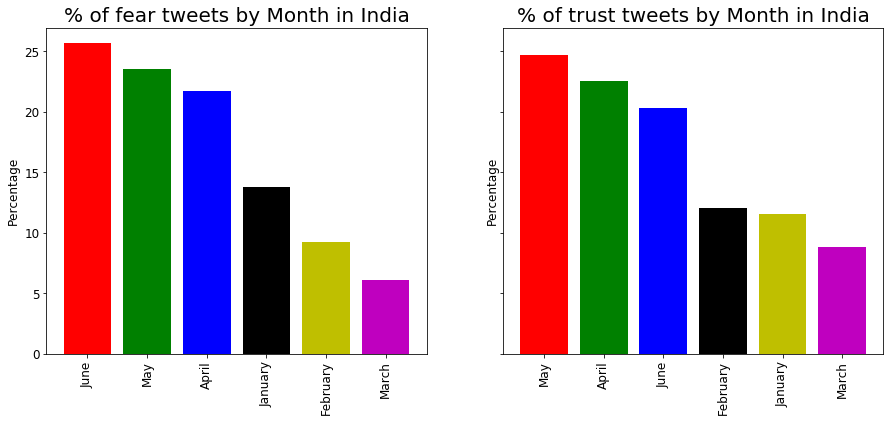}
      \caption{Percentage of Trust and Fear emotion in India per month}
      \label{figurelabel15}
   \end{figure}
Fig. 15 shows that the percentage of trust tweets are greater in May with the lockdown being imposed multiple times in this month, stricter restrictions, and division of cities into red, orange, and green zones. 
The trust percentage has reduced from May to June which might be the result of lockdown restrictions being lifted in this period.

\subsubsection{Brazil}
The number of cases in Brazil saw an exponential increase in the months after April as seen in Fig 16.
\begin{figure}[H]
      \centering
      \includegraphics[scale=0.2]{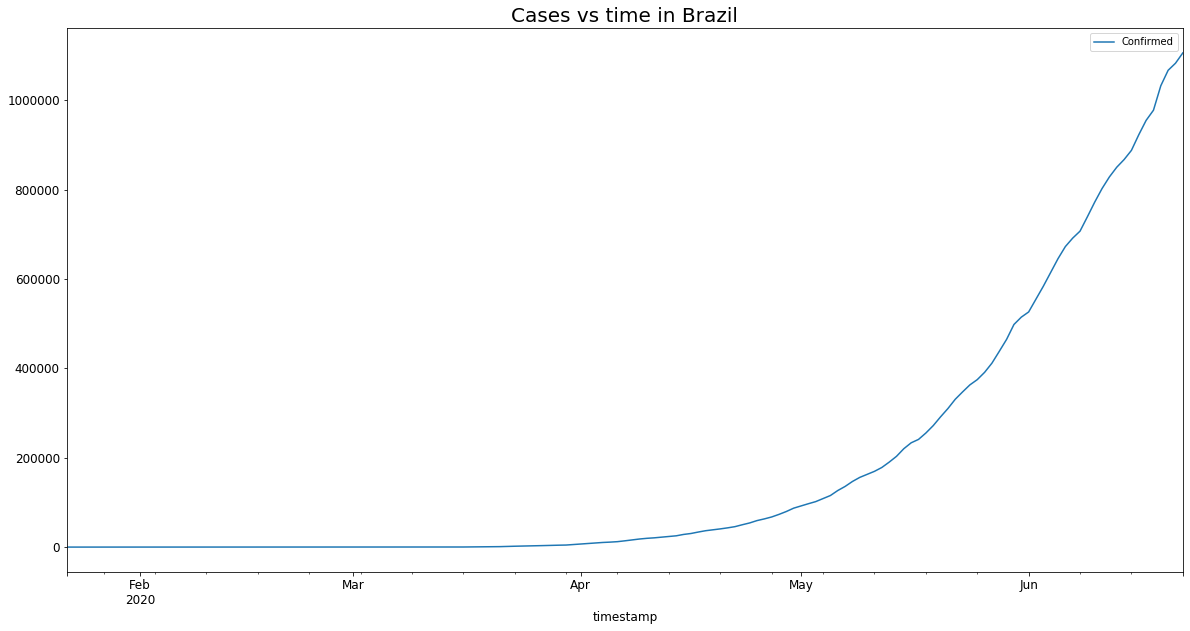}
      \caption{Number of cases in Brazil vs time}
      \label{figurelabel16}
   \end{figure}  
   
\begin{figure}[H]
      \centering
      \includegraphics[scale=0.25]{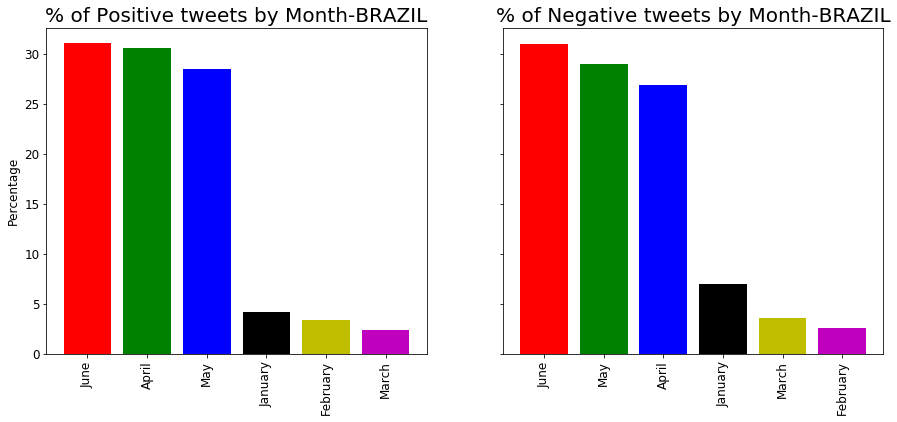}
      \caption{Percentage of positive and negative sentiment in Brazil per month}
      \label{figurelabel17}
   \end{figure}
 
\begin{figure}[H]
      \centering
      \includegraphics[scale=0.25]{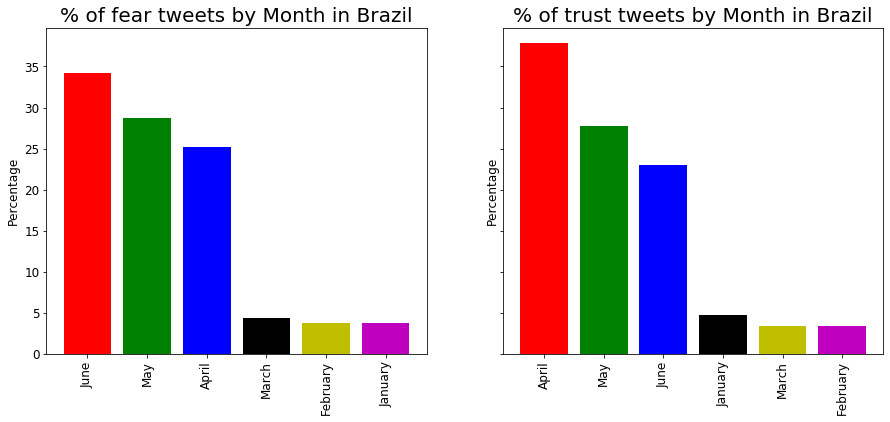}
      \caption{Percentage of Trust and Fear emotion in Brazil per month}
      \label{figurelabel18}
   \end{figure}  
The government of Brazil began providing financial assistance to the people in April, which could possibly explain the high percentage of trust emotion in April as shown in Fig. 18. In addition to financial assistance, 14000 people were discharged from a COVID ward after being declared as recovered[18].
On 23rd May, Brazil is declared as the 2nd worst-hit country, surpassing Russia. 
\subsection{Work From Home (WFH) and Online Learning (OL)}

\begin{figure}[H]
      \centering
      \includegraphics[scale=0.2]{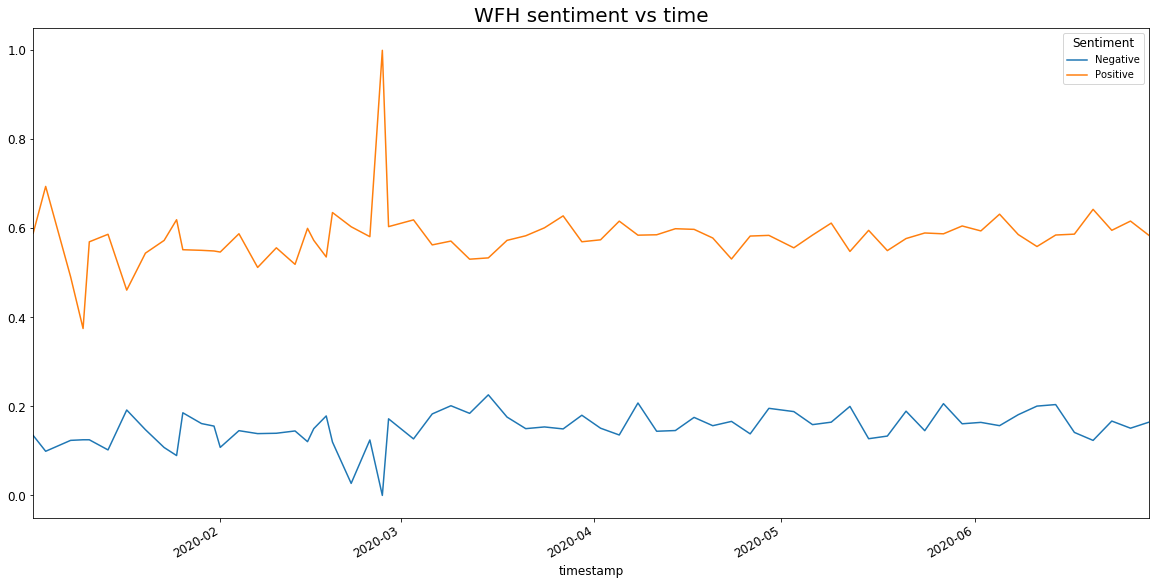}
      \caption{WFH sentiment vs time}
      \label{figurelabel19}
   \end{figure}   
The general positive sentiment towards Work From Home has consistently been higher than negative sentiment as shown in Fig. 19.
  
\begin{figure}[H]
      \centering
      \includegraphics[scale=0.2]{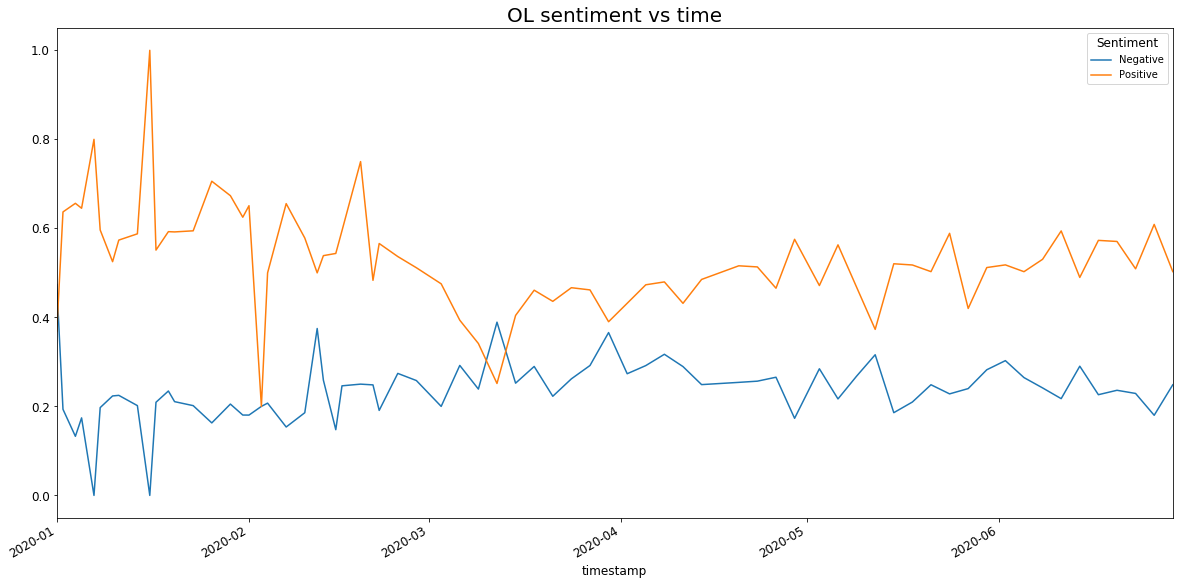}
      \caption{OL sentiment vs time}
      \label{figurelabel20}
   \end{figure}
The proportion of positive tweets have been greater for online learning. However, there have been instances where the proportion of negative sentiment has been greater than or equal to the proportion of positive sentiment as evidenced by Fig. 20.

\subsection{Conclusions}
Three datasets were created (covid19 tweets, WFH tweets, OL tweets) and exploratory data analysis was performed. This data was visualized in contrast to the Kaggle dataset having the number of confirmed cases in each country from the onset of the pandemic till June. The analysis of the created datasets was done with respect to the positive and negative sentiments, fear, and trust emotions exhibited by the tweets. These results were observed for countries that are worst hit by the pandemic and some interesting inferences were drawn. Two deep learning methods(LSTM and ANN) were used on the datasets for Sentiment Classification.
\newline Future work would involve the analysis of the sentiment of people towards other topics such as healthcare facilities, government response to the pandemic, offline examinations, mental health, etc. 

\addtolength{\textheight}{-12cm}   





\begin{thebibliography}{99}
\bibitem{c1}Rajput, Nikhil Kumar, Bhavya Ahuja Grover, and Vipin Kumar Rathi. "Word frequency and sentiment analysis of twitter messages during Coronavirus pandemic." arXiv preprint arXiv:2004.03925 (2020).
\bibitem{c2}M. Munikar, S. Shakya and A. Shrestha, "Fine-grained Sentiment Classification using BERT," ArXiv, 2019
\bibitem{c3}C. J. H. E. Gilbert, “Vader: A parsimonious rule-based model for
sentiment analysis of social media text,” in Eighth International
Conference on Weblogs and Social Media (ICWSM-14). Available at
(20/04/16)
http://comp.social.gatech.edu/papers/icwsm14.vader.hutto.pdf, 2014
\bibitem{c4}Hochreiter, S., Schmidhuber, J., “Long Short-Term Memory”,
Neural Computation 9 (8), 1997, pp. 1735–1780
\bibitem{5}Sundermeyer, Martin, Ralf Schlüter, and Hermann Ney. "LSTM neural networks for language modeling." Thirteenth annual conference of the international speech communication association. 2012.
\bibitem{6}Cannady, James. "Artificial neural networks for misuse detection." National information systems security conference. Vol. 26. 1998.
\bibitem{7}Samuel, Jim, et al. "Covid-19 public sentiment insights and machine learning for tweets classification." Information 11.6 (2020): 314.
\bibitem{8}Barkur, Gopalkrishna, and Giridhar B. Kamath Vibha. "Sentiment analysis of nationwide lockdown due to COVID 19 outbreak: Evidence from India." Asian journal of psychiatry (2020).
\bibitem{9}Li, Sijia, et al. "The impact of COVID-19 epidemic declaration on psychological consequences: a study on active Weibo users." International journal of environmental research and public health 17.6 (2020): 2032.
\bibitem{10}Jang, Hyeju, et al. "Exploratory analysis of covid-19 related tweets in north america to inform public health institutes." arXiv preprint arXiv:2007.02452 (2020).
\bibitem{11}Zhou, Jianlong, et al. "Examination of community sentiment dynamics due to covid-19 pandemic: a case study from Australia." arXiv preprint arXiv:2006.12185 (2020).
\bibitem{12}Taspinar, GitHub repository, https://github.com/taspinar/twitterscraper
\bibitem{13} World Cities Dataset - https://datahub.io/core/world-cities 
\bibitem{14} WHutto, C.J. \& Gilbert, Eric. (2015). VADER: A Parsimonious Rule-based Model for Sentiment Analysis of Social Media Text. Proceedings of the 8th International Conference on Weblogs and Social Media, ICWSM 2014. 
\bibitem{15} Covid19 Kaggle Dataset https://www.kaggle.com/imdevskp/corona-virus-report
\bibitem{16} Mohammad, Saif \& Turney, Peter. (2013). NRC emotion lexicon. 10.4224/21270984. 
\bibitem{17}Kingma, Diederik P., and J. Adam Ba. "A method for stochastic optimization. arXiv 2014." arXiv preprint arXiv:1412.6980 434 (2019).
\bibitem{18}Wikipedia. https://en.wikipedia.org/wiki/COVID-19\_pandemic\_in\_Brazil




\end{thebibliography}
\end{document}